\documentclass{article}
\usepackage[utf8]{inputenc}
\usepackage[margin=2cm]{geometry}
\usepackage{graphicx}
\usepackage{subfig}
\usepackage{url}
\usepackage{amsmath,amsfonts,amsthm}
\usepackage{amsmath,amsfonts,amssymb,amsthm}
\usepackage{authblk}
\usepackage{bm}
\usepackage{algorithm}
\usepackage{algpseudocode} 
\usepackage{titling}
\usepackage[most,skins,theorems]{tcolorbox}
\tcbset{
  aibox/.style={
    width=\linewidth,
    top=7pt,
    bottom=2pt,
    colback=blue!6!white,
    colframe=black,
    colbacktitle=black,
    enhanced,
    center,
    attach boxed title to top left={yshift=-0.1in,xshift=0.15in},
    boxed title style={boxrule=0pt,colframe=white,},
  }
}
\newtcolorbox{AIbox}[2][]{aibox,title=#2,#1}

\RequirePackage{mathpazo} 

\usepackage{tikz}
\usetikzlibrary{shapes,arrows.meta,positioning}

\tikzset{
    block/.style={
        rectangle,
        draw,        
        align=center,
        rounded corners,
        minimum height=2em,
        font=\sffamily
    },
    arrow/.style={
        -{Stealth[scale=1.2]},
        thick
    }
}

\title{\normalfont\bfseries\fontsize{16}{20} PCL-Reasoner-V1.5: Advancing Math Reasoning with \\ Offline Reinforcement Learning \vspace{-0.8cm}}

\author{%
\begin{minipage}{\textwidth}
\centering
    Yao Lu\textsuperscript{1,*}, 
    Dengdong Fan\textsuperscript{1,*}, 
    Jianzheng Nie\textsuperscript{1,*},
    Fan Xu\textsuperscript{1} \\ \vspace{-0.3cm}
    Jie Chen\textsuperscript{1,2,$\dagger$},
    Bin Zhou\textsuperscript{1,$\dagger$},
    Yonghong Tian\textsuperscript{1,2,$\dagger$}
\end{minipage}
}

\date{
\vspace{-0.2cm}
\textsuperscript{1}Peng Cheng Laboratory, \textsuperscript{2}Peking University 
}

\begin{document}

\maketitle

\begingroup
\renewcommand{\thefootnote}{} 
\footnotetext{\textsuperscript{*}Equal contribution. \textsuperscript{$\dagger$}Corresponding authors. Email: \texttt{\{luy01,chenj,zhoub03,tianyh\}@pcl.ac.cn}}
\endgroup

\vspace{-0.5cm}

\begin{abstract}
We present PCL-Reasoner-V1.5, a 32-billion-parameter large language model (LLM) for mathematical reasoning. The model is built upon Qwen2.5-32B and refined via  supervised fine-tuning (SFT) followed by reinforcement learning (RL). A central innovation is our proposed offline RL method, which provides superior training stability and efficiency over standard online RL methods such as GRPO. Our model achieves state-of-the-art performance among models post-trained on Qwen2.5-32B, attaining average accuracies of 90.9\% on AIME 2024 and 85.6\% on AIME 2025. Our work demonstrates offline RL as a stable and efficient paradigm for advancing reasoning in LLMs.
All experiments were conducted on Huawei Ascend 910C NPUs. 
\end{abstract}

\def\huggingface{\raisebox{-1.5pt}{\includegraphics[height=1.05em]{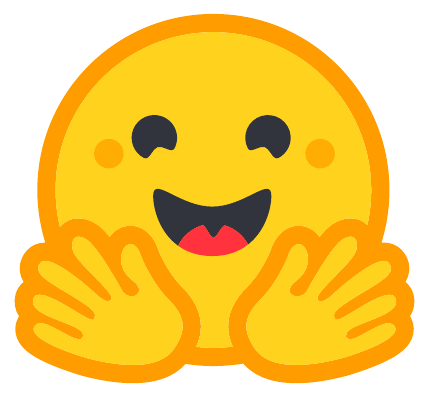}}}
\def\github{\raisebox{-1.5pt}{\includegraphics[height=1.05em]{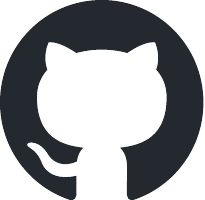}}}

\begin{center}
\begin{tabular}{rl}
\huggingface \hspace{-0.2cm} &  \url{https://huggingface.co/PCL-Reasoner/V1.5}\\
\github \hspace{-0.2cm} & \url{https://github.com/PCL-Reasoner/V1.5}\\
\end{tabular}
\end{center}

\vspace{-0.6cm}

\begin{figure}[h!]
\centering
\includegraphics[width=0.8\textwidth]{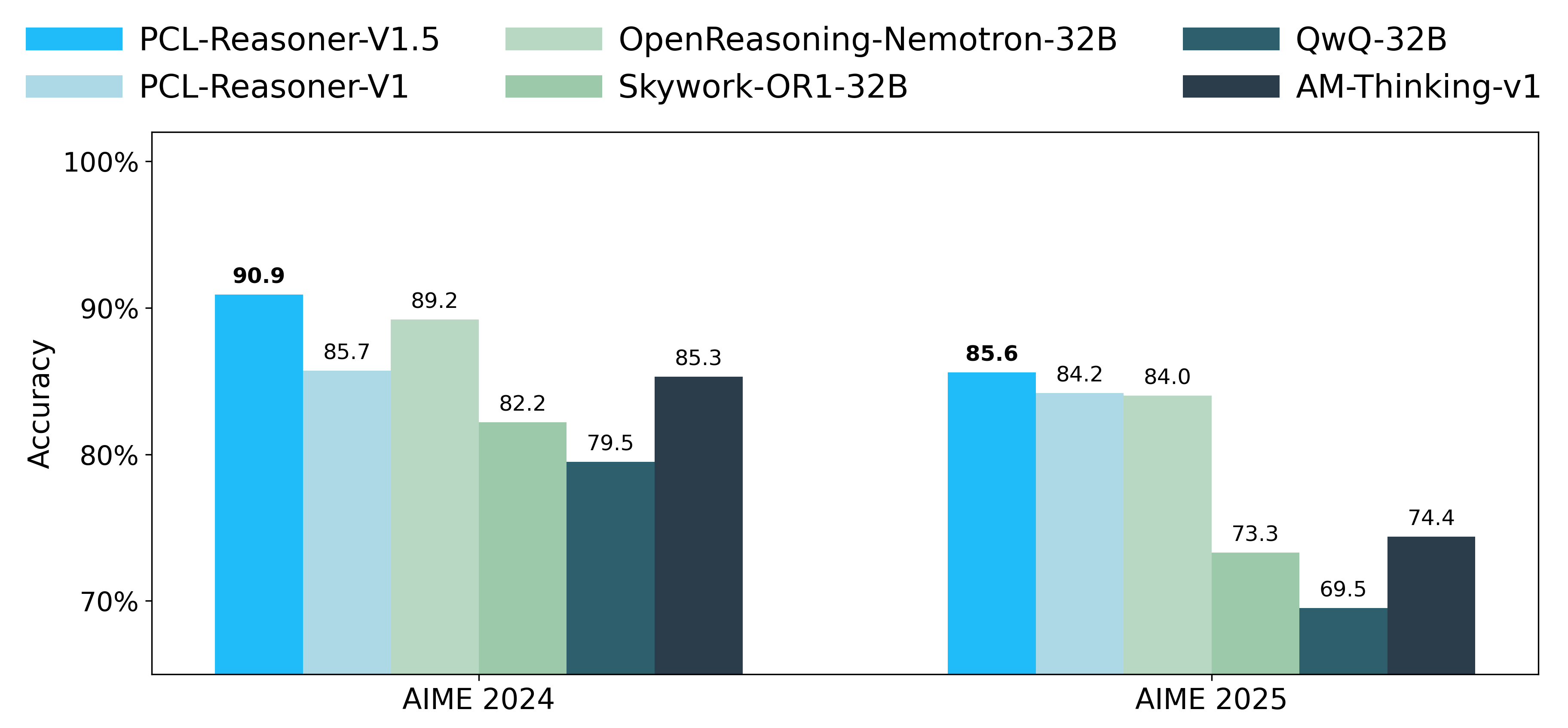}
\caption{Evaluation results of 32B models on AIME (pass@1)}
\end{figure}

\section{Introduction}

The rapid evolution of large language models (LLMs) has fundamentally expanded the frontiers of artificial intelligence, most notably in complex domains requiring systematic, multi-step reasoning \cite{brown2020language,openai2023gpt}. Among these, mathematical reasoning represents a particularly rigorous benchmark, as it demands precise logical deduction, symbolic manipulation, and structured problem-solving—capabilities that remain a significant challenge for even advanced models \cite{cobbe2021training,hendrycks2021measuring}.

A pivotal breakthrough in this domain is DeepSeek-R1 \cite{deepseek2025r1}, which demonstrates that superior reasoning performance can be elicited through the synergy of Chain-of-Thought (CoT)  \cite{wei2022chain} and Reinforcement Learning (RL). However, the prevailing paradigm relies predominantly on online RL approaches, such as Group Relative Policy Optimization (GRPO) \cite{shao2024deepseekmath}, which introduce significant technical challenges such as training instability and inefficiency (see Section 2).

To address these challenges, we introduce PCL-Reasoner-V1.5, a 32-billion-parameter LLM for mathematical reasoning. Built upon Qwen2.5-32B, our model undergoes supervised fine-tuning (SFT) followed by a novel offline RL approach. The offline RL approach circumvents the instabilities and computational costs associated with standard online RL while effectively enhancing reasoning performance.
PCL-Reasoner-V1.5 establishes a new state-of-the-art for models post-trained on Qwen2.5-32B, achieving average accuracies of 90.9\% on AIME 2024 and 85.6\% on AIME 2025. These results demonstrate that advanced reasoning capabilities can be achieved via offline RL, offering a more stable and computationally efficient alternative to online approaches.

All our experiments were conducted on Huawei Ascend 910C NPUs. We have released our model, data and code (including data processing, training, inference and evaluation) for the reproduction of our results.

\section{Online vs. Offline RL}

\begin{figure}[h!]
\vspace{0.5cm}
    \centering
        \subfloat[Online RL]{%
    \begin{tikzpicture}[
        node distance=0.5cm and 0.5cm, 
        auto
    ]

    \tikzset{every node/.style={text width=2.5cm, align=center}}
    
    \node[block, fill=blue!20,text width=1.5cm] (model0) {$\pi_{\theta_0}(y|x)$};
    \node[block, fill=green!20, text width=1.5cm, above=of model0] (infer1) {Inference};    
    \node[block, fill=white, text width=0.5cm, above=of infer1] (X1) {$\mathcal{X}_1$};
    \node[block, fill=white, text width=0.5cm, right=of infer1, xshift=0.1cm] (Y1) {$\mathcal{Y}_1$};    
    \node[block, fill=green!20, text width=1.5cm, right=of model0] (train1) {Training};
    \node[block, fill=blue!20, text width=1.5cm, right=of train1] (model1) {$\pi_{\theta_1}(y|x)$};

    \node[block, fill=green!20, text width=1.5cm, above=of model1] (infer2) {Inference};    
    \node[block, fill=white, text width=0.5cm, above=of infer2] (X2) {$\mathcal{X}_2$};
    \node[block, fill=white, text width=0.5cm, right=of infer2, xshift=0.1cm] (Y2) {$\mathcal{Y}_2$};    
    \node[block, fill=green!20, text width=1.5cm, right=of model1] (train2) {Training};
    \node[block, fill=blue!20, text width=1.5cm, right=of train2] (model2) {$\pi_{\theta_2}(y|x)$};

    \node[block, fill=green!20, text width=1.5cm, above=of model2] (infer3) {Inference};    
    \node[block, fill=white, text width=0.5cm, above=of infer3] (X3) {$\mathcal{X}_3$};
    \node[block, fill=white, text width=0.5cm, right=of infer3, xshift=0.1cm] (Y3) {$\mathcal{Y}_3$};
    \node[block, fill=green!20, text width=1.5cm, right=of model2] (train3) {Training};
    \node[block, fill=blue!20, text width=1.5cm, right=of train3] (model3) {$\pi_{\theta_3}(y|x)$};    
    
    \coordinate[right=of X1,xshift=1.8cm] (p1) {};
    \coordinate[right=of X2,xshift=1.8cm] (p2) {};    
    \coordinate[right=of X3,xshift=1.8cm] (p3) {};        
    
    \draw[arrow] (model0.north) -- (infer1.south);
    \draw[arrow] (infer1.east) -- (Y1.west);
    \draw[arrow] (train1.east) -- (model1.west);
    \draw[arrow] (X1.east) -- (p1) -- (p1|-train1.north);
    \draw[arrow] (X1.south) -- (infer1.north);
	\draw[arrow] (Y1.south) -- (Y1.south|-train1.north);
    \draw[arrow] (model0.east) -- (train1.west);
    
    \draw[arrow] (model1.north) -- (infer2.south);
    \draw[arrow] (infer2.east) -- (Y2.west);
    \draw[arrow] (train2.east) -- (model2.west);
    \draw[arrow] (X2.east) -- (p2) -- (p2|-train2.north);
    \draw[arrow] (X2.south) -- (infer2.north);
	\draw[arrow] (Y2.south) -- (Y2.south|-train2.north);
    \draw[arrow] (model1.east) -- (train2.west);    
    
    \draw[arrow] (model2.north) -- (infer3.south);
    \draw[arrow] (infer3.east) -- (Y3.west);
    \draw[arrow] (train3.east) -- (model3.west);
    \draw[arrow] (X3.east) -- (p3) -- (p3|-train3.north);
    \draw[arrow] (X3.south) -- (infer3.north);
	\draw[arrow] (Y3.south) -- (Y3.south|-train3.north);
    \draw[arrow] (model2.east) -- (train3.west);      
    \end{tikzpicture}}

    \vspace{0.5cm}
    
    \subfloat[Offline RL]{%
    \begin{tikzpicture}[
        node distance=0.5cm and 0.8cm, 
        auto
    ]

    \tikzset{every node/.style={text width=2.5cm, align=center}}
    
    \node[block, fill=blue!20,text width=1.5cm] (model0) {$\pi_{\theta_0}(y|x)$};
    \node[block, fill=green!20, text width=1.5cm, above=of model0] (infer1) {Inference};    
    \node[block, fill=white, text width=1.5cm, above=of infer1] (X1) {$\mathcal{X}_1,\mathcal{X}_2,\mathcal{X}_3$};

    \node[block, fill=green!20, text width=1.5cm, right=of model0,xshift=0.75cm] (train1) {Training};
    \node[block, fill=white, text width=1.5cm, right=of infer1,xshift=0.2cm] (Y1) {$\mathcal{Y}_1,\mathcal{Y}_2,\mathcal{Y}_3$};        
    \node[block, fill=blue!20, text width=1.5cm, right=of train1,xshift=0.75cm] (model1) {$\pi_{\theta_1}(y|x)$};

    \coordinate[right=of X1,xshift=2.15cm] (p1) {};
    
    \draw[arrow] (model0.north) -- (infer1.south);
    \draw[arrow] (infer1.east) -- (Y1.west);
    \draw[arrow] (train1.east) -- (model1.west);
    \draw[arrow] (X1.east) -- (p1) -- (p1|-train1.north);
    \draw[arrow] (X1.south) -- (infer1.north);
	\draw[arrow] (Y1.south) -- (Y1.south|-train1.north);
    \draw[arrow] (model0.east) -- (train1.west);
    
    \end{tikzpicture}}
    
    \vspace{0.2cm}
    \caption{Comparison between online and offline RL. $\mathcal{X}_i$ denotes a mini-batch of questions and $\mathcal{Y}_i$ denotes the corresponding model-inferred answers. (a) Online RL: An iterative cycle where $\mathcal{Y}_i$ is inferred on-the-fly by the current policy $\pi_{\theta_i}$ and used for immediate training. (b) Offline RL: A sequential process where all answers for the whole question set are inferred once, forming a static dataset for subsequent training.}
    \label{fig:rl}
\end{figure}
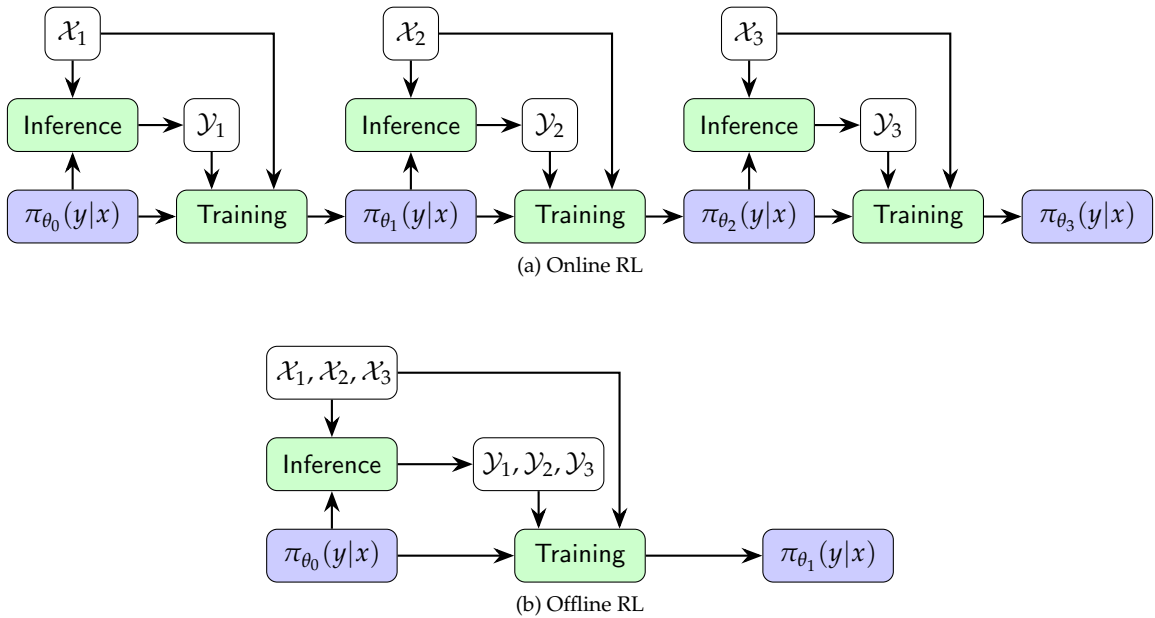

The prevailing RL paradigm for improving the reasoning capabilities of LLMs relies on online methods such as GRPO and its variants \cite{shao2024deepseekmath,deepseek2025r1,yu2025dapo,he2025skywork,zheng2025group}. This approach is inherently iterative: in each step, a mini-batch of data is sampled from the whole question dataset (with ground-truth), the model infers the answers given the mini-batch questions via sampling-based decoding, and a verifier assigns rewards to these answers given the corresponding ground-truth. The model is then trained using an RL loss. This closed-loop process is illustrated in Figure \ref{fig:rl}(a).

In contrast, offline RL methods, such as rejection sampling fine-tuning \cite{dong2023raft,yuan2023scaling}, operates on a fixed, pre-collected dataset. The whole  question set is first processed in a single pass to infer a static repository of answers and rewards. Subsequently, the model is trained on this fixed dataset without further inference during the training phase. This decoupled, single-stage process is shown in Figure \ref{fig:rl}(b).

Intuitively, online RL appears superior for reasoning tasks because the model dynamically infers new answers based on its updated parameters, benefiting from a continuous cycle of self-improvement and exploration that offline methods lack. In this work, we challenge this conventional wisdom and argue that offline RL is a strong alternative for fine-tuning reasoning LLMs.

\newpage

We posit that offline RL offers several distinct \textbf{advantages} over online RL.

\begin{itemize}
    \item \textbf{Training Stability.} Online RL is notoriously sensitive, often suffering from reward collapse or divergence due to the dynamic interplay between training and inference, a phenomenon exacerbated by numerical precision issues \cite{liu2025speed,qi2025defeating}. Offline RL mitigates this by decoupling inference from training. By training policy model on a fixed static dataset, the feedback loops that trigger instability are eliminated, resulting in a more robust and predictable training strategy.

    \item \textbf{Computational Efficiency.} In RL, the inference phase often becomes the computational bottleneck. Offline RL yields substantially higher inference throughput than typical online RL orchestration for two key reasons. First, by decoupling inference from the training loop, we can eliminate the substantial latency overhead caused by frequent model weight synchronization and cross-node communication. Second, decoupling enables inference to be deployed independently across heterogeneous compute nodes. This independence allows high-performance frameworks such as vLLM\footnote{\url{https://github.com/vllm-project/vllm}} to implement continuous batching efficiently: instead of waiting for synchronized batch completions across the entire cluster, each node can dynamically schedule new requests as previous ones finish. This maximizes hardware utilization and maintains a high-throughput pipeline for the training phase.

    \item \textbf{Engineering Simplicity.} Online RL necessitates complex orchestration frameworks, such as verl\footnote{\url{https://github.com/volcengine/verl}} and OpenRLHF\footnote{\url{https://github.com/OpenRLHF/OpenRLHF}}, to synchronize real-time rollouts with distributed gradient updates. Offline RL simplifies the pipeline into two sequential phases. This allows practitioners to optimize each stage independently using inference frameworks—such as vLLM for high-throughput inference and training frameworks such as Megatron for large-scale training, without the need for intricate orchestration.

    \item \textbf{Lower Experimentation Cost.} Training strategy exploration such as hyperparameter tuning in online RL is expensive, as every configuration change necessitates re-running the inference-training cycle. Offline RL allows for a ``infer once, train many'' workflow. After the inference phase, a single static dataset can be stored on disk and reused for training to ablate loss functions, learning rates, and other configurations.
\end{itemize}

However, offline RL also presents several \textbf{disadvantages}:

\begin{itemize}
    \item \textbf{Bounded Performance.} Offline RL is generally limited by the quality of the best examples within its static dataset. Unlike the "self-correction" or "emergence" seen in R1-Zero \cite{deepseek2025r1}, it cannot iteratively discover increasingly complex reasoning paths. Consequently, offline RL is best suited for refining models that already possess a strong performance baseline (e.g. after SFT).
    
    \item \textbf{Distribution Mismatch.} The primary theoretical concern in offline RL is the divergence between the inference (sampling) policy and the evolving training policy, which can lead to out-of-distribution (OOD) errors \cite{levine2020offline}. However, we argue this might be less critical for fine-tuning LLMs for two reasons. First, RL is typically applied to SFT models, which already possess strong performance. Therefore, 
    only a small fraction of the parameter space needs to tuned in the RL phase~\cite{mukherjee2025reinforcement}, keeping the distribution mismatch relatively small. Second, empirical evidence suggests that importance sampling, the standard remedy for distribution mismatch, is often ineffective or unnecessary in LLM RL settings \cite{wang2025aspo}.
\end{itemize}

While one could bridge the two paradigms by considering hybrid approaches such as iterative rejection sampling and iterative DPO \cite{pang2024iterative, lanchantin2025bridging}, we demonstrate in the next section that offline RL alone can lead to competitive results on challenging reasoning benchmarks.

\section{Our Approach}

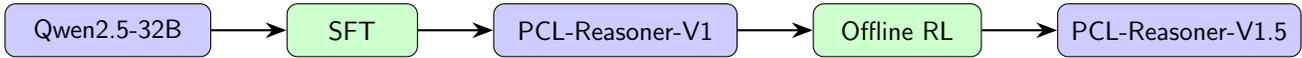
\begin{figure}[h!]
    \centering
    \begin{tikzpicture}[
        node distance=2cm and 1.0cm, 
        auto
    ]

    \tikzset{every node/.style={text width=2.5cm, align=center}}

    \node[block, fill=blue!20] (base) {Qwen2.5-32B};
    \node[block, fill=green!20, text width=1.5cm, right=of base] (sft) {SFT};
    \node[block, fill=blue!20, text width=3cm, right=of sft] (v1) {PCL-Reasoner-V1};
    \node[block, fill=green!20, text width=2cm, right=of v1] (orl) {Offline RL};
    \node[block, fill=blue!20, text width=3cm, right=of orl] (v15) {PCL-Reasoner-V1.5};

    \draw[arrow] (base.east) -- (sft.west);
    \draw[arrow] (sft.east) -- (v1.west);
    \draw[arrow] (v1.east) -- (orl.west);
    \draw[arrow] (orl.east) -- (v15.west);

    \end{tikzpicture}
    \caption{Training pipeline}
    \label{fig:pipeline}
\end{figure}

Our model, PCL-Reasoner-V1.5, is built upon Qwen2.5-32B~\footnote{\url{https://huggingface.co/Qwen/Qwen2.5-32B}}. We first applied SFT using CoT reasoning data distilled from DeepSeek-R1-0528~\footnote{\url{https://huggingface.co/deepseek-ai/DeepSeek-R1-0528}}, producing an intermediate model, PCL-Reasoner-V1, which already achieves high baseline performance. We then enhanced this model through an offline RL stage. The two-stage pipeline is depicted in Figure \ref{fig:pipeline}.

\newpage

\subsection{Prompt Template}

We adopt the following prompt template for reasoning, inspired by \cite{qwq32b,ji2025thinking}. We use the same prompt
template for SFT, RL and evaluation.
\begin{tcolorbox}[colframe=black,boxrule=1pt,boxsep=2pt,top=3pt,bottom=3pt,left=2pt,right=2pt]
\begin{verbatim}
<|im_start|>system
You are a helpful assistant. To answer the user's question, you first think about the 
reasoning process and then provide the user with the answer. The reasoning process and answer 
are enclosed within <think> </think> and <answer> </answer> tags, respectively, i.e., <think> 
reasoning process here </think> <answer> answer here </answer>.
<|im_start|>user
{input question} 
Please reason step by step, and put your final answer within \\boxed{}.<|im_end|>
<|im_start|>assistant
<think>
\end{verbatim}
\end{tcolorbox}

\subsection{SFT}

\paragraph{Data.} Our SFT training data was drawn from the mathematical subset of AM-DeepSeek-R1-0528-Distilled\footnote{\url{https://huggingface.co/datasets/a-m-team/AM-DeepSeek-R1-0528-Distilled}}. We conducted a data contamination test on the AIME 2024 and 2025 evaluation set and found no contamination. Training samples with answer sequences exceeding 32K tokens in length were removed. This filtering process yielded a final effective training set of 666K samples.

\paragraph{Training.} The global batch size was set to 128. The learning rate was set to $6\times10^{-5}$ and decayed to $1\times10^{-7}$ following a cosine schedule, with a warm-up ratio of 0.05. The momentum parameters in the AdamW optimizer were configured as $\beta_{1}=0.9$ and $\beta_{2}=0.95$. The entire training process spanned 4 epochs. To maximize computational efficiency, we enabled data packing during training. This feature allowed samples of varying lengths within each batch to be concatenated and integrated into the predefined sequence length of 32K tokens. By merging multiple short sequences into longer ones, it effectively eliminated redundant computations caused by sequence padding and significantly accelerated the training process.

\subsection{Offline RL.}

\paragraph{Data.} Our offline RL question set was derived from the mathematical subset of the Nemotron-Post-Training-Dataset-v1\footnote{\url{https://huggingface.co/datasets/nvidia/Nemotron-Post-Training-Dataset-v1}}. This dataset originally contained eight candidate answers per question inferred by DeepSeek-R1-0528, retaining only those that were correct. To isolate high-difficulty tasks suitable for eliciting complex reasoning, we applied two primary filtering heuristics: we excluded questions where (1) the average sequence length of the inferred answers was shorter than 32K tokens, and (2) all eight initial inferred answers were correct. This selection process ensured that our training data consisted of non-trivial questions that necessitated extensive reasoning chains. In the end, we obtained 6,068 difficult questions with ground-truth. We released this dataset to the public\footnote{\url{https://huggingface.co/PCL-Reasoner/V1.5}}.

\paragraph{Inference.} For every question in the RL dataset, we inferrer eight candidate answers using the SFT model, PCL-Reasoner-V1. We employed a sampling-based decoding strategy (top-$k$=40, top-$p$=0.95, and temperature=0.6) with the vLLM Ascend framework\footnote{\url{https://github.com/vllm-project/vllm-ascend}} on Huawei Ascend 910C NPUs. Each 32B model instance was deployed on a single compute node equipped with 8 NPUs using tensor parallelism of degree 8. 

\vspace{0.5cm}

\paragraph{Verification \& Filtering.}
We employed Qwen3-32B\footnote{\url{https://huggingface.co/Qwen/Qwen3-32B}} to verify the correctness of each inferred answer against the ground truth. A binary reward was assigned to each sample: $R=1$ for correct answers and $R=-1$ for incorrect ones. This process yielded a dataset of triplets, $\mathcal{D} = \{(x_i, y_i, R_i)\}_{i=1}^N$, where $x_i$ represents the question, $y_i$ the inferred answer, and $R_i$ the associated reward.
To improve training, we applied two filters. First, we excluded questions where the rewards lacked variance (i.e., where all eight answers were entirely correct or entirely incorrect), mirroring the internal mechanism of GRPO~\cite{xiong2025minimalist}. Second, to prevent the model from learning a length bias, we discarded any positive samples whose answer exceeds 48K tokens. The final refined dataset contains $N=30,215$ samples, comprising 14,512 positive samples ($R_i=1$) and 15,703 negative samples ($R_i=-1$).

\paragraph{Training.}

Given  $\mathcal{D} = \{(x_i, y_i, R_i)\}_{i=1}^N$, we trained the policy model $\pi_{\theta}(y|x)$  to minimize the RL loss 
\begin{align}
L(\theta) =   -\sum_i  R_i \pi_{\theta}(y_i|x_i).
\label{eq:loss}
\end{align}
Inspired by \cite{zheng2025group}, we define the policy value $\pi_{\theta}(y_i|x_i)$ as the geometric mean of token-level probabilities:
\begin{align}
\pi_{\theta}(y_i|x_i) = \Big(\prod_{t=1}^{|y_i|} p_{\theta}(y_{i,t} | x_i, y_{i,<t})\Big)^{1/|y_i|} 
                     = \exp\!\Big(\frac{1}{|y_i|}\sum_{t=1}^{|y_i|}\log p_{\theta}(y_{i,t} | x_i, y_{i,<t})\Big),
\end{align}
where $p_{\theta}(y_{i,t}|x_i, y_{i,<t})$ denotes the probability of the $t$-th token in $y_i$ given the prefix $y_{i,<t}$, and $|y_i|$ is the total token count of answer $y_i$. We did not apply importance sampling, which was shown to be ineffective in LLM RL training \cite{wang2025aspo}, even though the answers were generated from distribution other than the current policy model.

We trained for 800 steps with a global batch size of 128. Optimization was performed using AdamW with $\beta_{1}=0.9$, $\beta_{2}=0.95$ and a weight decay constant of 0.1. The learning rate followed a cosine schedule, starting at $1\times10^{-6}$ and decaying to $1\times10^{-7}$, with no warm-up phase. To preserve numerical precision, we trained in FP16 format rather than BF16, a choice informed by \cite{qi2025defeating}.
The training infrastructure consisted of 8 compute nodes, each equipped with 8 Huawei Ascend 910C NPUs. We adapted the MindSpeed-LLM framework\footnote{\url{https://gitcode.com/Ascend/MindSpeed-LLM}}  for training and scaled the model using tensor parallelism of degree 8 and pipeline parallelism of degree 4. Memory constraints were mitigated via activation recomputation and optimizer state swapping.

\begin{figure}[h!]
\centering
\includegraphics[width=0.6\textwidth]{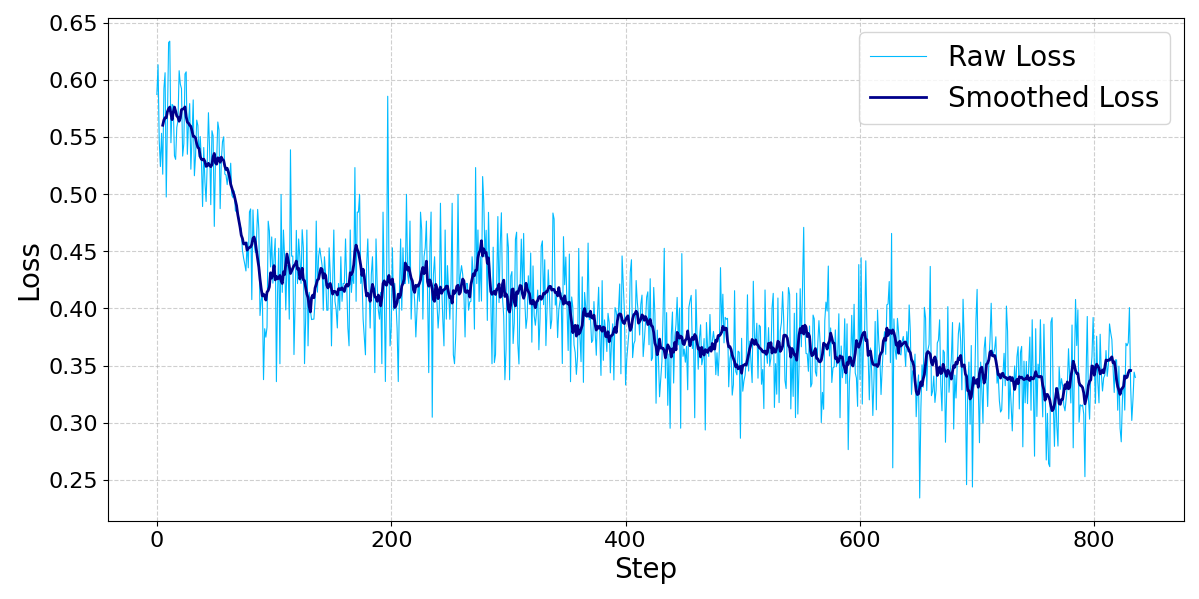}
\caption{Training loss $L_{\text{norm}}(\theta)$}
\label{fig:train_loss}
\end{figure}

To facilitate a clearer visualization of the training trajectory, we plot the following normalized loss in Figure \ref{fig:train_loss}:
\begin{align}
L_{\text{norm}}(\theta) = \sum_i \Big[ \frac{R_i+1}{2} (1-\pi_{\theta}(y_i|x_i)) + \Big(1-\frac{R_i+1}{2}\Big)\pi_{\theta}(y_i|x_i) \Big].
\end{align}
This function is non-negative and produces the same gradient as the original objective in Eq. (\ref{eq:loss}), given $R_i = \pm 1$.

\vspace{0.2cm}

\newpage

\section{Evaluation}

We evaluated our trained models on the AIME 2024 and 2025 benchmarks, each consisting of 30 Olympic-level math problems. For inference, we configured the decoding parameters with a temperature of 0.6, top-$k$=40, top-$p$=0.95, and a maximum token limit of 129,024. Each question was evaluated over 32 sampled responses.
The results are presented in Table 1. Our model, PCL-Reasoner-V1.5, achieves state-of-the-art performance among all models derived from the Qwen2.5-32B.

\begin{table}[h!]
\centering
\begin{tabular}{|l|c|c|}
\hline
Model & AIME 2024 & AIME 2025 \\
\hline
DeepSeek-R1 & 79.8 & 70 \\
DeepSeek-R1-0528 & 91.4 & 87.5 \\
Qwen3-235B-A22B & 85.7 & 81.5 \\
OpenAI-o3	& \textbf{91.6} & \textbf{88.9} \\
Gemini-2.5-Pro-0506	& 90.8 & 83 \\
\hline
QwQ-32B &	79.5	& 69.5 \\
DeepSeek-R1-Distill-Qwen-32B	& 72.6	& 49.6 \\
Skywork-OR1-32B	& 82.2 &	73.3 \\
AM-Thinking-v1	& 85.3 & 74.4 \\
OpenReasoning-Nemotron & 89.2 & 84.0 \\
PCL-Reasoner-V1	& {85.7} & {84.2} \\
PCL-Reasoner-V1.5	& \textbf{90.9} & \textbf{85.6} \\
\hline
\end{tabular}
\caption{Evaluation results (pass@1)}
\end{table}

To understand the source of the improvement from offline RL training, we analyzed the models' response (answer) length. As shown in Table \ref{table:response}, RL training leads to a substantial increase in the average response length on both benchmarks. This behavioral shift suggests that RL encourages the model to engage in more extensive and deliberate reasoning, potentially by learning to avoid past errors and to explore solution paths more thoroughly.

\begin{table}[h!]
\centering
\begin{tabular}{|l|c|c|}
\hline
Model & AIME 2024 & AIME 2025 \\
\hline
PCL-Reasoner-V1	& 22,535 & 25,993 \\
PCL-Reasoner-V1.5 &  30,560  & 39,835 \\
\hline
\end{tabular}
\caption{Average response length}
\label{table:response}
\end{table}

\begin{figure}[h!]
\centering
\includegraphics[width=0.75\textwidth]{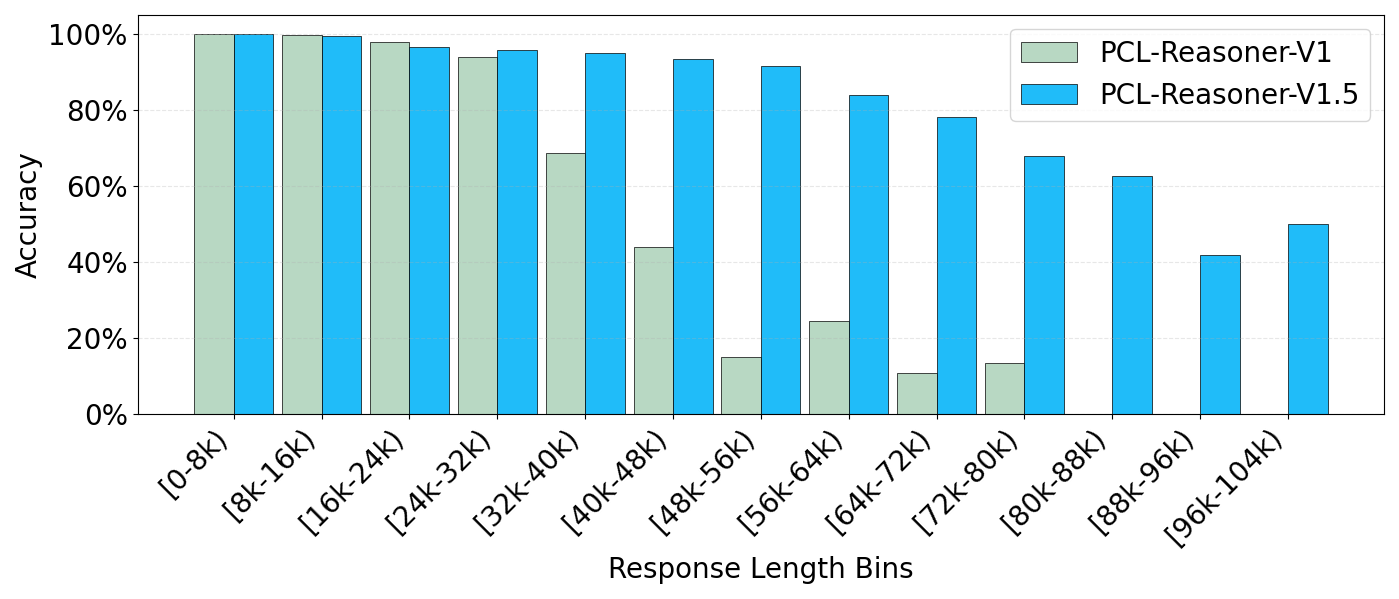}
\caption{Accuracy on each response length range on AIME 2024 \& 2025}
\label{fig:acc_len}
\end{figure}

To further pinpoint the nature of this improvement, we disaggregate accuracy by response length of the AIME 2024 and 2025 evaluation responses in Figure \ref{fig:acc_len}. The SFT model (PCL-Reasoner-V1) performs poorly on problems requiring long CoTs ($\geq$32 K). In contrast, the RL-trained model shows a dramatic improvement in accuracy specifically on questions requiring long-CoT reasoning, demonstrating that RL effectively enhances the model's capacity for sustained, complex reasoning.

\vspace{0.2cm}

\newpage

\section{Conclusion}

In this work, we introduced PCL-Reasoner-V1.5, a 32-billion-parameter LLM for  mathematical reasoning. By building upon Qwen2.5-32B and employing a two-stage pipeline—SFT followed by a novel offline RL approach—we have demonstrated that high-tier reasoning capabilities can be achieved.

Empirical evaluations on the AIME 2024 and 2025 benchmarks confirm the efficacy of our approach. PCL-Reasoner-V1.5 achieved state-of-the-art performance among models post-trained on Qwen2.5-32B, with accuracies of 90.9\% and 85.6\%, respectively. Our analysis reveals that the primary driver of this improvement is the model's enhanced long-CoT reasoning elicited by RL.

Our findings challenge the prevailing reliance on online methods like GRPO. We have shown that the offline RL paradigm offers significant advantages in training stability, computational efficiency, and engineering simplicity, making it a strong alternative to online RL for fine-tuning reasoning LLMs. Future work will explore hybrid iterative approaches to further push the boundaries of reasoning capabilities in LLMs.

\bibliographystyle{apalike}
\bibliography{ref}

\end{document}